\documentclass[a4paper, 10pt, conference]{ieeeconf} 

\IEEEoverridecommandlockouts                              

\usepackage{graphics} 
\usepackage{epsfig} 
\usepackage{times} 
\usepackage{amsmath} 
\usepackage{amssymb}  
\usepackage[T1]{fontenc}

\title{\LARGE \bf
Homography matrix based trajectory planning method for robot uncalibrated visual servoing
}

\author{Zhongtao Fu$^{1,3}$, $\;$ Xiaoyu Lei$^{1}$, $\;$ Xubing Chen$^{1}$, $\;$ Mohamed Ibrahim Ahmed$^{1}$, $\;$Cong Zhang$^{1}$, $\;$ Miao Li$^{2}$, $\;$Tao Huang$^{3}$
\thanks{*This work is supported by the Natural Science Foundation of China (51805380, 51875415), the Foundation of State Key Laboratory of Digital Manufacturing Equipment and Technology (DMETKF2022019), the International Exchanges 2021 Cost Share (RS-NSFC) award (IECNSFC211345), the Central Guidance on Local Science and Technology Development Foundation of Hubei Province (2019ZYYD010), the Graduate Education Innovation Foundation of Wuhan Institute of Technology(CX2022079).(Corresponding author: Zhongtao Fu,  hustfzt@gmail.com).}
\thanks{$^{1}$Xiaoyu Lei, Zhongtao Fu, Xubing Chen, Cong ZHANG are with the School of Mechanical and Electrical Engineering, Wuhan Institute of Technology, Wuhan 430205, China.}%
\thanks{$^{2}$Miao Li is with the Institute of Technological Sciences, Wuhan University, Wuhan 430072, China.}
\thanks{$^{3}$Zhongtao Fu, Tao Huang are with the State Key Laboratory of Digital Manufacturing Equipment and Technology, Huazhong University of Science and Technology, Wuhan 430704, China.}%
}

\begin{document}

\maketitle
\begin{abstract}

In view of the classical visual servoing trajectory planning method which only considers the camera trajectory, this paper proposes one homography matrix based trajectory planning method for robot uncalibrated visual servoing. Taking the robot-end-effector frame as one generic case, eigenvalue decomposition is utilized to calculate the infinite homography matrix of the robot-end-effector trajectory, and then the image feature-point trajectories corresponding to the camera rotation is obtained, while the image feature-point trajectories corresponding to the camera translation is obtained by the homography matrix. According to the additional image corresponding to the robot-end-effector rotation, the relationship between the robot-end-effector rotation and the variation of the image feature-points is obtained, and then the expression of the image trajectories corresponding to the optimal robot-end-effector trajectories (the rotation trajectory of the minimum geodesic and the linear translation trajectory) are obtained. Finally, the optimal image trajectories of the uncalibrated visual servoing controller is modified to track the image trajectories. Simulation experiments show that, compared with the classical IBUVS method, the proposed trajectory planning method can obtain the shortest path of any frame and complete the robot visual servoing task with large initial pose deviation.

\end{abstract}

\section{INTRODUCTION}

The perceptive feedback of visual sensors in the various scenarios is employed to control robot, such as grasping planning, welding-seam tracking, ultrasonic scanning [1-3]. Classical visual servoing method typically require the calibration procedure, while the accuracy of the calibration results has a significant impact on control performance, as well as the issues of cumbersome and time-consuming calibration process [4-5]. Consequently, the uncalibrated visual servoing method is widely used in practical robot control.

Visual servoing methods are classified into three types based on the input signals: position-based visual servoing (PBVS), hybrid visual servoing (HVS), and image-based visual servoing (IBVS).  PBVS and HVS are not suitable for uncalibrated visual servoing tasks because of the required 3D reconstruction. Because the IBVS method only uses image information for control, it has attracted many researchers' interest on robot uncalibrated visual servoings. The image-based uncalibrated visual servoing (IBUVS) method designs the proportional controller using the inverse matrix of the image Jacobian, so estimating the image Jacobian matrix is the key to the IBUVS method, which is generally estimated by the nonlinear optimization methods (Gauss-Newton method, Levenberg-Marquard method, or Kalman filtering method) [6-7]. In recent years, depth learning method [8-9] has also been applied to uncalibrated visual servoing tasks, and the higher precision image Jacobian matrix is obtained by constructing a network model [10-11]. Although the image Jacobian matrix accuracy met the actual control requirements, the IBUVS method itself has some drawbacks [12]: 1) The IBUVS method is locally stable, and can ensure convergence only when the initial pose is close to the desired pose. 2) In some cases, the image Jacobian matrix may be singular. 3) The IBUVS method cannot guarantee the robot's optimal trajectory.

In order to solve the aforementioned issues pertaining to the existing IBUVS method, trajectory planning can be performed during the robot visual servoing process [13]. One method is to divide the trajectory into multiple small trajectories for control by using dynamic planning [14], or to obtain rotation and translation motion of the shortest translation and rotation path by decomposing the homography matrix [15-16]. The basic matrix can also be used for optimal trajectory planning [17]. However, in the process of decomposition, at least a rough understanding of the camera's intrinsic parameters is required, and because multiple solutions exist, additional information is required to determine the true solution. Another approach is to plan directly in homography space and generate the trajectory using the homography matrix [18-19]. Gong et al. [20] used the homography matrix to generate the optimal camera trajectory. This method avoids decomposing the homography matrix and obtains the homography matrix without scaling ambiguity using additional image information. These methods only consider the camera trajectory in 3D space and do not take into general situations, such as the motion trajectory of the robot end-effector. As a result, this paper proposes an improved robot uncalibrated visual servoing trajectory planning method. Using the robot end-effector frame as an example, the trajectory of image feature-point is planned using the homography matrix and the infinite homography matrix, which corresponds to the motion trajectory of the robot end-effector. The visual servoing controller is then designed to follow the planned trajectory, ensuring that the robot's end-effector moves to the desired pose in the shortest path.

\section{Robot uncalibrated visual servoing system}

\begin{figure}[!t]
\centering
\includegraphics[width=0.45\textwidth]{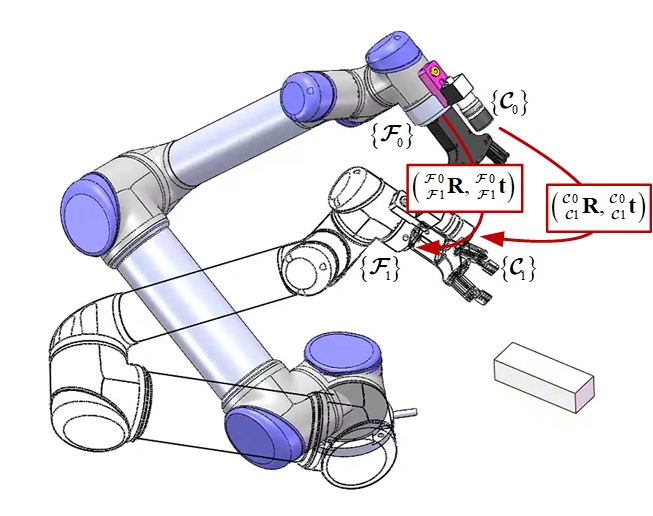}
\caption{Robot uncalibrated visual servoing system.}
\label{fig1}
\end{figure}

Fig.1 depicts the robot uncalibrated visual servoing system model, in which the subscripts 0 and 1 of the camera frame $\left\{ \mathcal{C} \right\}$ and the robot end-effector frame $\left\{ \mathcal{F} \right\}$, represent the current and desired pose, and the transformation relationship between them is $\left( {}_{\mathcal{C}1}^{\mathcal{C}0}\mathbf{R},{}_{\mathcal{C}1}^{\mathcal{C}0}\mathbf{t} \right)$ and $\left( {}_{\mathcal{F}1}^{\mathcal{F}0}\mathbf{R},{}_{\mathcal{F}1}^{\mathcal{F}0}\mathbf{t} \right)$. It is unnecessary to calibrate the hand-eye relationship of the robot and the camera intrinsic parameters in advance. The task of the visual servoing is to move the current robot end-effector pose $\left\{ {{\mathcal{F}}_{0}} \right\}$ to the desired pose $\left\{ {{\mathcal{F}}_{1}} \right\}$.

\subsection{Imaging model of monocular camera}

Imaging model of monocular camera and the homography matrix are extremely crucial in robot uncalibrated visual servoing. As shown in Fig.2, the point $\mathcal{X}$ is expressed respectively in the current and desired camera frames as ${}_{{}}^{0}\mathcal{X}={{\left[ {}_{{}}^{0}x,{}_{{}}^{0}y,{}_{{}}^{0}z \right]}^{T}}$ and ${}_{{}}^{1}\mathcal{X}={{\left[ {}_{{}}^{1}x,{}_{{}}^{1}y,{}_{{}}^{1}z \right]}^{T}}$, the corresponding pixel coordinates are ${}_{{}}^{0}\mathcal{P}={{\left[ {}_{{}}^{0}u,{}_{{}}^{0}v,1 \right]}^{T}}$ and ${}_{{}}^{1}\mathcal{P}={{\left[ {}_{{}}^{1}u,{}_{{}}^{1}v,1 \right]}^{T}}$. The following relationships can be derived by means of the coordinate transformation and the imaging model of monocular camera:

$$
\left\{
\begin{array}{c}
   {}_{{}}^{1}z{}_{{}}^{1}\mathcal{P}=K{}_{{}}^{1}\mathcal{X}  \\
   {}_{{}}^{0}z{}_{{}}^{0}\mathcal{P}=K\left( {}_{\mathcal{C}1}^{\mathcal{C}0}\mathbf{R}\cdot {}_{{}}^{1}\mathcal{X}+{}_{\mathcal{C}1}^{\mathcal{C}0}\mathbf{t} \right)  \\
\end{array}
\right.
\eqno{(1)}
$$
where ${}_{\mathcal{C}1}^{\mathcal{C}0}\mathbf{R}\in \mathbb{S}\mathbb{O}\left( 3 \right)$ and ${}_{\mathcal{C}1}^{\mathcal{C}0}\mathbf{t}\in {{\mathbb{R}}^{3}}$ are the rotation matrix and translation vector of camera frame $\left\{ {{\mathcal{C}}_{1}} \right\}$ relative to $\left\{ {{\mathcal{C}}_{0}} \right\}$; $\mathbf{K}\in {{\mathbb{R}}^{3\times 3}}$ represents the camera intrinsic parameter matrix.

The homography matrix represents the mapping between the image feature-point pairs (${}_{{}}^{0}\mathcal{P},{}_{{}}^{1}\mathcal{P}$) on two image planes with different view field, considering the plane constraint ${}_{{}}^{1}n{}_{{}}^{1}\mathcal{X}+{}_{{}}^{1}d=0$, ${}_{{}}^{0}\mathcal{P}$ can be expressed as:
$$
{}_{{}}^{0}\mathcal{P}=\frac{{}_{{}}^{1}z}{{}_{{}}^{0}z}{}_{1}^{0}\mathbf{H}{}_{{}}^{1}\mathcal{P}=\frac{{}_{{}}^{1}z}{{}_{{}}^{0}z}\mathbf{K}({}_{\mathcal{C}1}^{\mathcal{C}0}\mathbf{R}-\frac{{}_{\mathcal{C}1}^{\mathcal{C}0}\mathbf{t}}{{}_{{}}^{1}d}{}_{{}}^{1}\mathbf{n}){{\mathbf{K}}^{-1}}{}_{{}}^{1}\mathcal{P}
\eqno{(2)}
$$
where ${}_{1}^{0}\mathbf{H}\in {{\mathbb{R}}^{3\times 3}}$ is a homography matrix, which can be solved by Direct
Linear Transformation (DLT) technique [21]. When ${}_{{}}^{1}d$ tends to infinity, the infinite homography matrix ${}_{1}^{0}{{\mathbf{H}}^{\infty }}$ is expressed as:
$$
 {}_{1}^{0}{{\mathbf{H}}^{\infty }}=\mathbf{K}{}_{\mathcal{C}1}^{\mathcal{C}0}\mathbf{R}{{\mathbf{K}}^{-1}}
 \eqno{(3)}
 $$
Four vanishing points are employed at least to estimate infinite homography matrix ${}_{1}^{0}{{\mathbf{H}}^{\infty }}$ according to the illustration [22-23].

\begin{figure}[!t]
\centering
\includegraphics[width=0.4\textwidth]{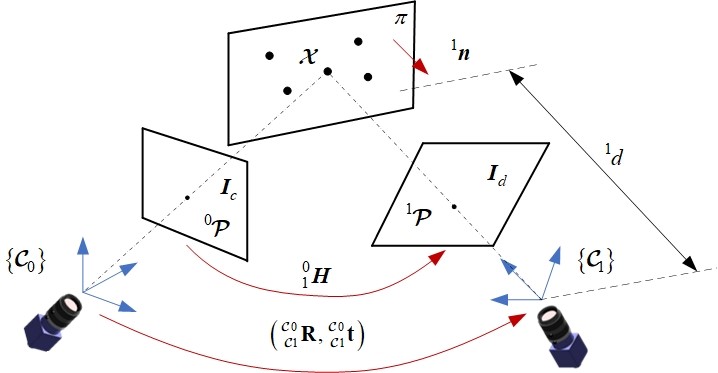}
\caption{Camera imaging model.}
\label{fig2}
\end{figure}

\subsection{Robot uncalibrated visual servoing approach}

Image-based robot uncalibrated visual servoing only use image to control the robot, and the trajectory of the image features needs to be converted into the motion of the end-effector, which is established by means of the image Jacobian matrix as:
$$\dot{y}=\mathbf{J}\dot{q} \eqno{(4)}$$
where $\mathbf{J}$ is the image Jacobian matrix, $\dot{y}\in {{\mathbf{R}}^{m}}$ is the image feature-point velocity vector, and $\dot{q}\in {{\mathbf{R}}^{n}}$ is the robot joint position velocity vector.

The controller design for uncalibrated visual servoing is to perform online estimation of the image Jacobian matrix without the calibrations of the robot hand-eye relationship or the camera intrinsic parameters [24-25]. And the proportional feedback controller based on the inverse matrix of the Jacobian matrix is designed to complete the control task. The control law is written as follows:
	$$\dot{q}=\lambda {{\mathbf{J}}^{+}}\dot{y} \eqno{(5)}$$
where $\lambda $ is the control gain and ${{\mathbf{J}}^{+}}$ is the pseudo-inverse matrix of $\mathbf{J}$.

\section{Trajectory planning with uncalibrated visual servoing}

For the robot visual servoing task, the image feature-point trajectory corresponding to the motion trajectory of the camera frame $\left\{ \mathcal{C} \right\}$ is firstly analyzed, and then extended to the general situation with the robot end-effector frame $\left\{ \mathcal{F} \right\}$, the image feature-point trajectory corresponding to its motion trajectory is analyzed sequentially.

\subsection{Trajectory planning of rigid-body motion} 

The rigid-body motion trajectory in 3D space consists of the rotational and translational motions. The rotation matrix ${}_{\mathcal{C}1}^{\mathcal{C}0}\mathbf{R}$ is equivalent to the form of axis angle $\theta \mathbf{k}$, where $\mathbf{k}$ is the rotation axis and $\theta $ is the rotation angle. It can be decomposed into eigenvalues, ${}_{\mathcal{C}1}^{\mathcal{C}0}\mathbf{R}=\mathbf{U}\cdot {}_{1}^{0}\Delta \cdot {{\mathbf{U}}^{-1}}$, where ${}_{1}^{0}\Delta $ is a diagonal matrix composed of three eigenvalues $\cos \theta +i\sin \theta ,\cos \theta -i\sin \theta ,1$, the eigenvalues are only related to the rotation angle $\theta $, and $\mathbf{U}$ is an identity orthogonal matrix composed of eigenvectors, and the eigenvectors are only related to the rotation axis $\mathbf{k}$. Because the rotation matrix ${}_{\mathcal{C}1}^{\mathcal{C}0}\mathbf{R}$ has a relationship with the rotation axis $\mathbf{k}$: ${}_{\mathcal{C}1}^{\mathcal{C}0}\mathbf{Rk}=\mathbf{k}$. As a result, for the best rotation motion, the rotation axis $\mathbf{k}$ is constant, i.e., $\mathbf{U}$ is constant, and only the eigenvalue of ${}_{\mathcal{C}1}^{\mathcal{C}0}\mathbf{R}$ is changing. The rotation angle about time is defined as: $\theta (t)=t\theta $, $t\in [0,1]$. So the expression of the rotation matrix ${}_{\mathcal{C}t}^{\mathcal{C}0}\mathbf{R}$ of the orientation at time $t$ relative to the initial orientation is:
$$
\begin{array}{c}
   {}_{\mathcal{C}t}^{\mathcal{C}0}\mathbf{R}=\mathbf{U}\cdot {}_{t}^{0}\Delta \cdot {{\mathbf{U}}^{-1}} \\ 
  {}_{t}^{0}\Delta = diag(\cos \theta (t)+i\sin \theta (t),\cos \theta (t)-i\sin \theta (t),1)  
\end{array}
\eqno{(6)}
$$

Linear interpolation is performed in 3D space to obtain the translation path. As a result, the translation vector ${}_{\mathcal{C}t}^{\mathcal{C}0}\mathbf{t}$ with respect to the initial position is expressed as:
	$${}_{\mathcal{C}t}^{\mathcal{C}0}\mathbf{t}=t{}_{\mathcal{C}1}^{\mathcal{C}0}\mathbf{t} \eqno{(7)}$$

\subsection{Camera motion trajectory}

According to (1), the relationship between the pose at time $t$ and the initial pose is expressed as:
$$
 \begin{array}{c}
   {}^{0}z{}^{0}\mathcal{P}={}^{t}z{}_{t}^{0}{{\mathbf{H}}^{\infty }}{}^{t}\mathcal{P}+{}_{\mathcal{C}t}^{\mathcal{C}0}\mathbf{T} \\ 
  {}_{t}^{0}{{\mathbf{H}}^{\infty }}=\mathbf{K}{}_{\mathcal{C}t}^{\mathcal{C}0}\mathbf{R}{{\mathbf{K}}^{-1}} \\ 
  {}_{\mathcal{C}t}^{\mathcal{C}0}\mathbf{T}=\mathbf{K}{}_{\mathcal{C}t}^{\mathcal{C}0}\mathbf{t} \\ 
\end{array}
\eqno{(8)}
$$
And the corresponding image feature-point trajectory is:
	$${}^{t}z{}^{t}\mathcal{P}={{({}_{t}^{0}{{\mathbf{H}}^{\infty }})}^{-1}}({}^{0}z{}^{0}\mathcal{P}-{}_{\mathcal{C}t}^{\mathcal{C}0}\mathbf{T}) \eqno{(9)}$$

Because the intrinsic parameter matrix $\mathbf{K}$ and the camera pose transformation matrix (${}_{\mathcal{C}t}^{\mathcal{C}0}\mathbf{R}$,${}_{\mathcal{C}t}^{\mathcal{C}0}\mathbf{t}$) are both unknown for the robot uncalibrated visual servoing task, the image feature-point trajectory cannot be obtained directly. But the infinite homography matrix ${}_{t}^{0}{{\mathbf{H}}^{\infty }}$ and rotation matrix $
{}_{\mathcal{C}t}^{\mathcal{C}0}\mathbf{R}$ are similar matrices, the expression of ${}_{t}^{0}{{\mathbf{H}}^{\infty }}$ can be obtained by combining (3) and (6):
$$
 {}_{t}^{0}{{\mathbf{H}}^{\infty }}=\mathbf{KU}\cdot {}_{t}^{0}\Delta \cdot {{(\mathbf{KU})}^{-1}}
 \eqno{(10)}
$$

As a result, the matrix ${}_{1}^{0}{{\mathbf{H}}^{\infty }}$ are decomposed to obtain the eigenvector matrix $\mathbf{KU}$, and the eigenvalues are interpolated to obtain the value of the matrix ${}_{t}^{0}{{\mathbf{H}}^{\infty }}$.

The expression of ${}_{\mathcal{C}t}^{\mathcal{C}0}\mathbf{T}$ corresponding to the camera translation trajectory can be obtained using (1), (2), and (7) as follows:
$$
{}_{\mathcal{C}t}^{\mathcal{C}0}\mathbf{T}=t\mathbf{K}{}_{\mathcal{C}1}^{\mathcal{C}0}\mathbf{t}=t{}_{\mathcal{C}1}^{\mathcal{C}0}\mathbf{T}=t{}^{1}z({}_{1}^{0}\mathbf{H}-{}_{1}^{0}{{\mathbf{H}}^{\infty }}){}^{1}\mathcal{P}  
\eqno{(11)}
$$

To obtain the trajectory of image feature-points, substitute (10) and (11) into (9):
$$
\begin{array}{c}
   {}^{t}z{}^{t}\mathcal{P}={{({}_{t}^{0}{{\mathbf{H}}^{\infty }})}^{-1}}({}^{0}z{}^{\mathcal{C}0}\mathcal{P}-t{}^{1}z({}_{1}^{0}\mathbf{H}-{}_{1}^{0}{{\mathbf{H}}^{\infty }}){}^{1}\mathcal{P}) \\ 
  ={}^{1}z{{({}_{t}^{0}{{\mathbf{H}}^{\infty }})}^{-1}}({}_{1}^{0}\mathbf{H}-t({}_{1}^{0}\mathbf{H}-{}_{1}^{0}{{\mathbf{H}}^{\infty }})){}^{1}\mathcal{P}  
\end{array}
\eqno{(12)}
$$
where ${}^{t}z$ and ${}^{1}z$ have no effect on the solution of the image feature trajectory due to the normalization of homogeneous coordinates.

\subsection{Robot end-effector motion trajectory}

The camera frame $\left\{ \mathcal{C} \right\}$ and the robot end-effector frame $\left\{ \mathcal{F} \right\}$ have the following relationship:
	$${}_{\mathcal{C}t}^{\mathcal{C}0}\mathbf{R}={}_{\mathcal{F}}^{\mathcal{C}}\mathbf{R}{}_{\mathcal{F}t}^{\mathcal{F}0}\mathbf{R}{}_{\mathcal{F}}^{\mathcal{C}}{{\mathbf{R}}^{-1}}
  \eqno{(13)}$$
	$${}_{\mathcal{C}t}^{\mathcal{C}0}\mathbf{t}={}_{\mathcal{F}}^{\mathcal{C}}\mathbf{R}({}_{\mathcal{F}t}^{\mathcal{F}0}\mathbf{R}-\mathbf{I}){}_{\mathcal{C}}^{\mathcal{F}}\mathbf{t}+{}_{\mathcal{F}}^{\mathcal{C}}\mathbf{R}{}_{\mathcal{F}t}^{\mathcal{F}0}\mathbf{t}
  \eqno{(14)}$$
where ${}_{\mathcal{C}}^{\mathcal{F}}\mathbf{R}$ and ${}_{\mathcal{C}}^{\mathcal{F}}\mathbf{t}$ are the rotation matrix and translation vector in the robot hand-eye relationship.

Combining (8), (13) and (14), the following expressions are obtained:
$$\begin{array}{c}
{}^{0}z{}^{0}\mathcal{P}={}^{t}z{}_{t}^{0}{{\mathbf{H}}^{\infty }}{}^{t}\mathcal{P}+\left( {}_{t}^{0}{{\mathbf{H}}^{\infty }}-\mathbf{I} \right){}_{\mathcal{C}}^{\mathcal{F}}\mathbf{T}+{}_{\mathcal{F}t}^{\mathcal{F}0}\mathbf{T} \\ 
{}_{t}^{0}{{\mathbf{H}}^{\infty }}=\mathbf{K}{}_{\mathcal{F}}^{\mathcal{C}}\mathbf{R}{}_{\mathcal{F}t}^{\mathcal{F}0}\mathbf{R}{{\left( \mathbf{K}{}_{\mathcal{F}}^{\mathcal{C}}\mathbf{R} \right)}^{-1}} \\ 
{}_{\mathcal{F}t}^{\mathcal{F}0}\mathbf{T}=\mathbf{K}{}_{\mathcal{F}}^{\mathcal{C}}\mathbf{R}{}_{\mathcal{F}t}^{\mathcal{F}0}\mathbf{t} \\ 
{}_{\mathcal{C}}^{\mathcal{F}}\mathbf{T}=\mathbf{K}{}_{\mathcal{F}}^{\mathcal{C}}\mathbf{R}{}_{\mathcal{C}}^{\mathcal{F}}\mathbf{t}  
\end{array}
\eqno{(15)}
$$
The corresponding image feature-point trajectory is as follows:
	$${}^{t}z{}^{t}\mathcal{P}={{\left( {}_{t}^{0}{{\mathbf{H}}^{\infty }} \right)}^{-1}}\left( {}^{0}z{}^{0}\mathcal{P}-\left( {}_{t}^{0}{{\mathbf{H}}^{\infty }}-\mathbf{I} \right){}_{\mathcal{C}}^{\mathcal{F}}\mathbf{T}-{}_{\mathcal{F}t}^{\mathcal{F}0}\mathbf{T} \right) \eqno{(16)}$$

Because ${}_{t}^{0}{{\mathbf{H}}^{\infty }}$ and ${}_{\mathcal{F}t}^{\mathcal{F}0}\mathbf{R}$ are similar matrices, the matrix ${}_{1}^{0}{{\mathbf{H}}^{\infty }}$ are decomposed to obtain the eigenvector matrix, and the eigenvalues are interpolated to obtain the value of matrix ${}_{t}^{0}{{\mathbf{H}}^{\infty }}$.
Assuming that the robot end-effector only rotates at time t, combining (2) and (15) obtains:
$$\left\{ \begin{array}{c}
     {}^{t}z{}^{t}\mathcal{P}={}^{0}z{}_{0}^{t}{{\mathbf{H}}_{r}}^{\infty }{}^{0}\mathcal{P}+\left( {}_{0}^{t}{{\mathbf{H}}_{r}}^{\infty }-\mathbf{I} \right){}_{\mathcal{C}}^{\mathcal{F}}\mathbf{T}  \\
     {}^{t}z{}^{t}\mathcal{P}={}^{0}z{}_{0}^{t}{{\mathbf{H}}_{r}}{}^{0}\mathcal{P}  \\
\end{array} \right.
\eqno{(17)}
$$
where ${}_{0}^{t}{{\mathbf{H}}_{r}}^{\infty }$ is the infinite homography matrix when only rotating, and ${}_{0}^{t}{{\mathbf{H}}_{r}}$ is the homography matrix when only rotating. As a result, ${}_{\mathcal{C}}^{\mathcal{F}}\mathbf{T}$ is expressed as follows:
	$$\frac{1}{{}^{1}z}{}_{\mathcal{C}}^{\mathcal{F}}\mathbf{T}={{\left( {}_{0}^{t}{{\mathbf{H}}_{r}}^{\infty }-\mathbf{I} \right)}^{+}}({}_{0}^{t}{{\mathbf{H}}_{r}}-{}_{0}^{t}{{\mathbf{H}}_{r}}^{\infty }){}_{1}^{0}\mathbf{H}{}^{1}\mathcal{P}
 \eqno{(18)}
 $$
where ${{\left( {}_{t}^{0}{{\mathbf{H}}_{r}}^{\infty }-\mathbf{I} \right)}^{+}}$ is the pseudo inverse of matrix $\left( {}_{t}^{0}{{\mathbf{H}}_{r}}^{\infty }-\mathbf{I} \right)$.

Combining (2), (15) and (18), the expression of the trajectory ${}_{\mathcal{F}t}^{\mathcal{F}0}\mathbf{T}$ can be obtained as:
$${}_{\mathcal{F}t}^{\mathcal{F}0}\mathbf{T}=t\cdot {}^{1}z\left( \left( {}_{1}^{0}\mathbf{H}-{}_{1}^{0}{{\mathbf{H}}^{\infty }} \right){}^{1}\mathcal{P}-\left( {}_{1}^{0}{{\mathbf{H}}^{\infty }}-\mathbf{I} \right)\frac{1}{{}^{1}z}{}_{\mathcal{C}}^{\mathcal{F}}\mathbf{T} \right)  \eqno{(19)}$$

Substituting (10), (18), and (19) into (16), the trajectory of the image feature-points corresponding to the robot end-effector trajectory is expressed as:
$$
\begin{array}{c}
{}^{t}z{}^{t}\mathcal{P}={}^{1}z{{\left( {}_{t}^{0}{{\mathbf{H}}^{\infty }} \right)}^{-1}}\left( \mathbf{A}(t){}^{1}\mathcal{P}+\mathbf{B}(t)\frac{1}{{}^{1}z}{}_{\mathcal{C}}^{\mathcal{F}}\mathbf{T} \right) \\ 
\mathbf{A}(t)={}_{1}^{0}\mathbf{H}-t\left( {}_{1}^{0}\mathbf{H}-{}_{1}^{0}{{\mathbf{H}}^{\infty }} \right) \\ 
\mathbf{B}(t)=t\left( {}_{1}^{0}{{\mathbf{H}}^{\infty }}-\mathbf{I} \right)-\left( {}_{t}^{0}{{\mathbf{H}}^{\infty }}-\mathbf{I} \right)  
\end{array}
\eqno{(20)}
$$
As a result, in order to obtain the image feature-point trajectory corresponding to the robot end-effector trajectory, the matrixs ${}_{1}^{0}\mathbf{H}$ and ${}_{1}^{0}{{\mathbf{H}}^{\infty }}$ are necessary, and a pure rotation motion on the robot end-effector frame $\left\{ \mathcal{F} \right\}$ must also be performed to obtain the initial estimated value of ${}_{\mathcal{C}}^{\mathcal{F}}\mathbf{T}$.

\section{VERIFICATION AND DISCUSSION}

\begin{figure}[!t]
\centering
\includegraphics[width=0.4\textwidth]{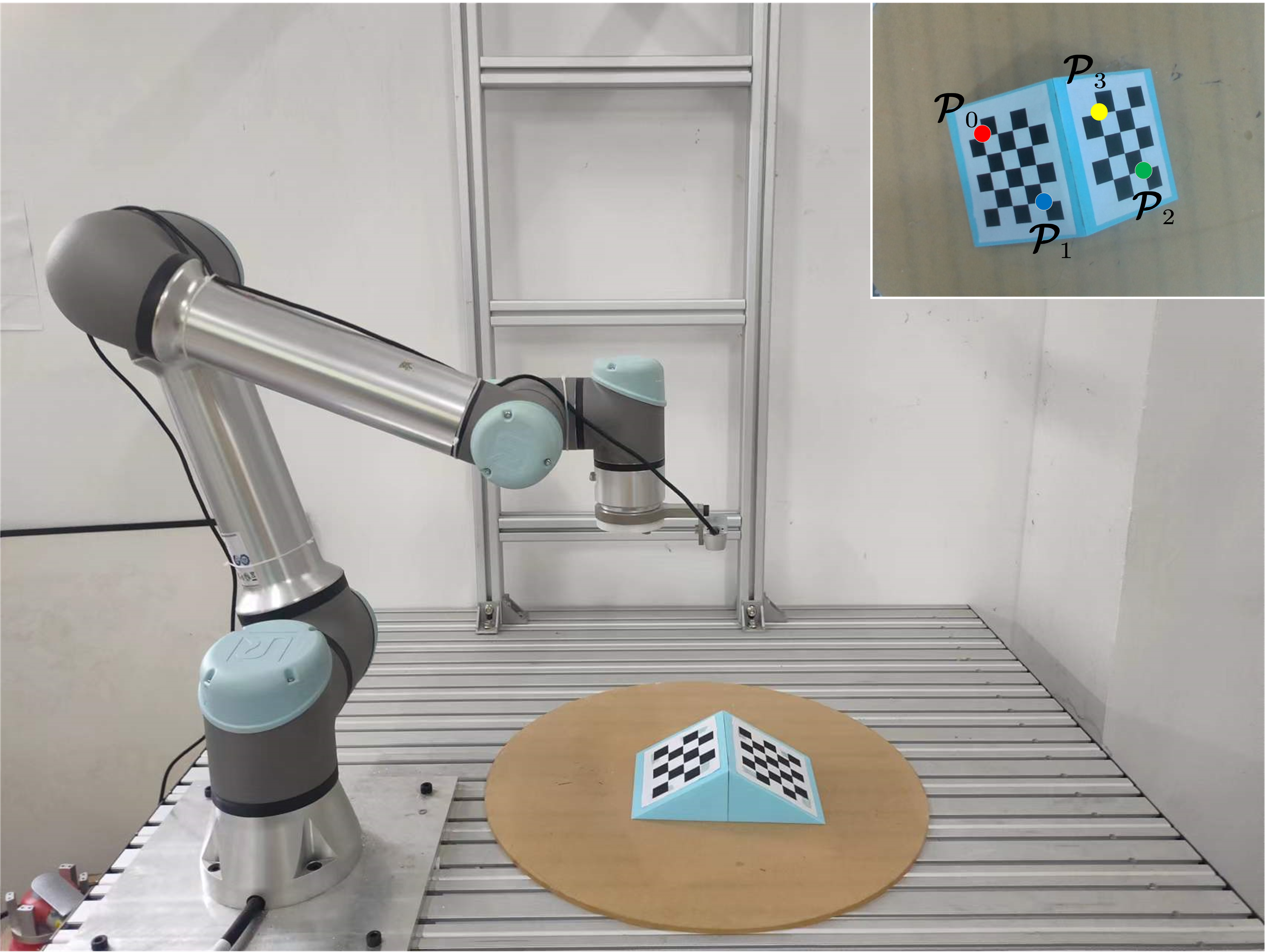}
\caption{Experimental setup.}
\label{fig3}
\end{figure}

To validate the robustness and efficacy of the proposed method for robot uncalibrated visual servoing, results are extracted based on a comparison with classical IBUVS method, while considering different initial poses of the robot in the experiment. The experimental analysis and comparison of the IBUVS and the proposed improved uncalibrated visual servoing trajectory planning method, where IBUVS-C represents the motion trajectory of the camera, and IBUVS-R represents the motion trajectory of the robot end-effector. 

Fig.3 shows the experimental setup, in which a RealSense camera is mounted on the robot end-effector and the target object is fixed on the workstation. The rotation error $er{{r}_{r}}$ and translation error $er{{r}_{t}}$ are calculated by means of (21) and (22) as:
$$
er{{r}_{t}}=\left\| {}^{0}\mathbf{t}-{}^{1}\mathbf{t} \right\| 
\eqno{(21)}
$$
$$
er{{r}_{r}}=\theta \left( {}_{\mathcal{F}1}^{\mathcal{F}0}\mathbf{R} \right) 
\eqno{(22)}
$$
where ${}^{0}\mathbf{t}$ and ${}^{1}\mathbf{t}$ represent the robot end-effector's current and desired position vectors; ${}_{\mathcal{F}1}^{\mathcal{F}0}\mathbf{R}$ represents the rotation matrix of the desired robot end-effector frame relative to the current frame; and $\theta $ represents the rotation angle.

\begin{figure*}[htbp]
	\centering
	\begin{minipage}{0.49\linewidth}
		\centering
		\includegraphics[width=1\linewidth]{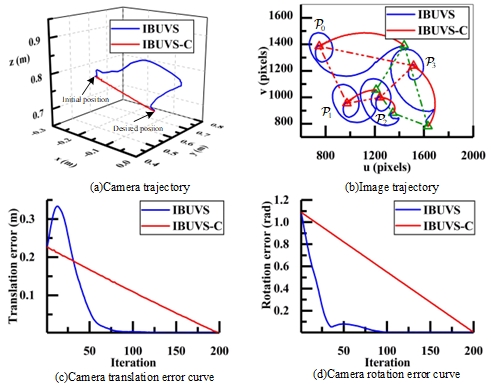}
		\caption{Comparison results of IBUVS and IBUVS-C.}
		\label{fig4}
	\end{minipage}
	\begin{minipage}{0.49\linewidth}
		\centering
		\includegraphics[width=1\linewidth]{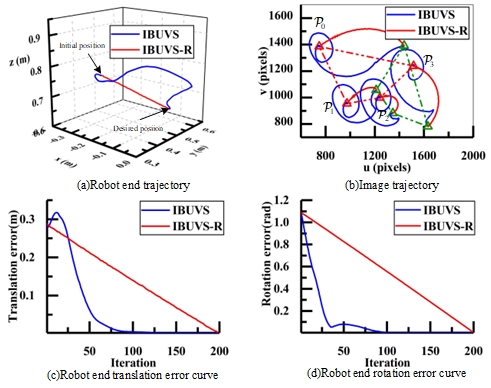}
		\caption{Comparison results of IBUVS and IBUVS-R.}
		\label{fig5}
	\end{minipage}
\end{figure*}

Case I: The initial pose of robot end-effector with respect to the desired end-effector frame is $\left[ \begin{matrix}
   {}_{\mathcal{F}1}^{\mathcal{F}0}\mathcal{X} & \Delta \theta \mathbf{k}  \\
\end{matrix} \right]=\left[ \begin{matrix}
   -250mm & -100mm & 100mm & 10{}^\circ  & -10{}^\circ  & 60{}^\circ   \\
\end{matrix} \right]$, ensuring that both the classical IBUVS and the method proposed in this paper can complete the robot uncalibrated visual servoing task. Figs. 4 and 5 show the comparison results calculated by the two methods. The camera motion and robot end-effector motion trajectories planed by the IBUVS method are both very long and discontinuous. But the trajectories planed by the IBUVS-C method are straight, which can better achieve the task of robot uncalibrated visual servoing trajectory planning and move to the desired pose with high accuracy. In order to quantitatively illustrate the advantages of the proposed method, the translation distance and rotation angle were selected as a metric for evaluating the two methods. In Fig.4, the camera translation distance and rotation angle obtained by the classical IBUVS method are 0.74m and 83°, while the IBUVS-C method yields 0.22m and 62°. In Figure 5, The robot end-effector translation distance and rotation angle obtained by the classical IBUVS method are 0.83m and 83°, whereas the robot end-effector translation distance and rotation angle obtained by the IBUVS-R method are 0.29m and 62°. Obviously, the motion trajectory of the camera and robot end-effector, as well as the motion trajectory of the image feature-points, have been optimized using the trajectory planning method proposed in this paper.

\begin{figure*}[htbp]
	\centering
	\begin{minipage}{0.49\linewidth}
		\centering
		\includegraphics[width=1\linewidth]{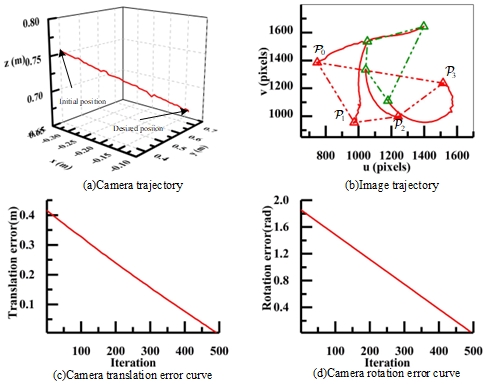}
		\caption{Calculated results of IBUVS-C.}
		\label{fig6}
	\end{minipage}
	\begin{minipage}{0.49\linewidth}
		\centering
		\includegraphics[width=1\linewidth]{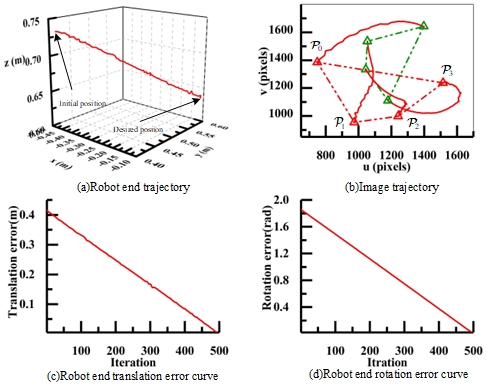}
		\caption{Calculated results of IBUVS-R.}
		\label{fig7}
	\end{minipage}
\end{figure*}

Case II: The initial pose of robot end-effector with respect to the desired end-effector frame is $\left[ \begin{matrix}
   {}_{\mathcal{F}1}^{\mathcal{F}0}\mathcal{X} & \Delta \theta \mathbf{k}  \\
\end{matrix} \right]=\left[ \begin{matrix}
   -350mm & -200mm & 100mm & 20{}^\circ  & -20{}^\circ  & 100{}^\circ   \\
\end{matrix} \right]$. The classical IBUVS method will not converge due to the large angle. At this time, the IBUVS-C and IBUVS-R are still valid, and the results are shown in Figs.6 and 7. It can be seen in Sec.III that IBUVS-C and IBUVS-R can complete the robot uncalibrated visual servoing task based on the motion trajectory expression of the camera and the robot end-effector. When there is a significant error between the initial and desired pose, the proposed trajectory planning method will deviate slightly due to the calculating error of the infinite homography matrix, while the overall trajectory is the shortest.

\section{CONCLUSIONS}

This paper presents one methodology for robot uncalibrated visual servoing trajectory planning based on the homography matrix. In terms of the generic case of the robot end-effector frame, the trajectory of image feature-points is generated, and the effectiveness of the proposed method is validated by means of the experimental data. The main contributions of this work can be synopsized as follows:

(1) The infinite homography matrix obtained by eigenvalue decomposition is employed to the optimal trajectory planning of the image feature-points corresponding to the camera and the robot end-effector, which can be applicable to any frame control in robot system.

(2) Experiments show that the proposed method can guarantee the convergence of the visual servoing task under large initial pose deviations. Compared to the classical IBUVS method, the proposed IBUVS-C and IBUVS-R methods can reduce the translation distance and rotation angle by 0.54m and 21°, respectively, required for the robot uncalibrated visual servoing task.

Furthermore, the proposed trajectory planning method is utilized mainly for the uncalibrated visual servoing issues of a single robot equipped with visual sensors, however, the the visual servoing task of multi-robots will be carried out in the future research.

\end{document}